# A Review of Computational Approaches for Evaluation of Rehabilitation Exercises

Yalin Liao[1], Aleksandar Vakanski[1*], Member, IEEE, Min Xian[1], Member, IEEE, David Paul[2], and Russell Baker[2]

[1] Department of Computer Science, University of Idaho, Idaho Falls, USA

[2] Department of Movement Sciences, University of Idaho, Moscow, USA

*Corresponding Author: 1776 Science Center Drive, Idaho Falls, ID 83402; vakanski@uidaho.edu; (+1)208-757-5422

*Abstract*— Recent advances in data analytics and computer-aided diagnostics stimulate the vision of patient-centric precision healthcare, where treatment plans are customized based on the health records and needs of every patient. In physical rehabilitation, the progress in machine learning and the advent of affordable and reliable motion capture sensors have been conducive to the development of approaches for automated assessment of patient performance and progress toward functional recovery. The presented study reviews computational approaches for evaluating patient performance in rehabilitation programs using motion capture systems. Such approaches will play an important role in supplementing traditional rehabilitation assessment performed by trained clinicians, and in assisting patients participating in home-based rehabilitation. The reviewed computational methods for exercise evaluation are grouped into three main categories: discrete movement score, rule-based, and template-based approaches. The review places an emphasis on the application of machine learning methods for movement evaluation in rehabilitation. Related work in the literature on data representation, feature engineering, movement segmentation, and scoring functions is presented. The study also reviews existing sensors for capturing rehabilitation movements and provides an informative listing of pertinent benchmark datasets. The significance of this paper is in being the first to provide a comprehensive review of computational methods for evaluation of patient performance in rehabilitation programs.

## 1. Introduction

Physical rehabilitation is commonly prescribed to benefit patients who suffer from physical impairments or disabilities, or need to restore functional abilities after injury or surgery [1]–[3]. Numerous studies in the literature underline the essential role of physical rehabilitation for improved patient outcomes and emphasize the strong correlation between exercise intensity and outcomes of rehabilitation programs [4]–[7]. However, rehabilitation treatment imposes a substantial economic burden on patients and healthcare systems [8]–[10]. For instance, the cost of physical rehabilitation programs in the US in 2007 was about 13.5 billion dollars based on the Medical Expenditure Panel Survey generated by the US federal government [9]. The expenditure was produced by nearly 9 million adults during approximately 88 million physical rehabilitation episodes.

In rehabilitation programs, a clinician instructs patients and monitors their performance of rehabilitation exercises in a clinical setting. This type of rehabilitation treatment is restricted by the availability of trained clinicians and it places demands on patients' schedules. To increase the flexibility of rehabilitation programs, home-based rehabilitation is often employed as a supplement to clinic-based programs. In home-based regimens, a clinician customizes a personal rehabilitation plan for a patient consisting of a set of recommended exercises. Patients then perform the exercises following the given instructions, record their daily progress in a logbook, and visit the clinic periodically for progress assessment. Reports in the literature indicate more than 90% of rehabilitation programs are executed in an at-home environment [11]. Nevertheless, a number of medical sources report low levels of patient motivation and adherence to the prescribed exercise regimens in home-based rehabilitation, leading to prolonged treatment duration and increased healthcare cost [12], [13]. Although many factors that reduce patient motivation and engagement in rehabilitation training have been identified, the lack of timely feedback and real-time supervision by a healthcare professional in an at-home setting is often cited as the most influential factor [14]. Poor motivation and supervision promote further risk because patients may perform exercises incorrectly as a result of those factors, which increases the risk of re-injury [3], [4], [12].

Accordingly, there is a demand for novel tools and equipment to support home-based rehabilitation, such as robotic assistive devices [15], exoskeletons, haptic devices [16], and virtual gaming environments [17]. With the advent of low-cost motion capture sensors, like Microsoft Kinect [18] and Asus Xtion [19], there has been a surge in related biomedical applications [20], [21]. VERA (Virtual Exercise Rehabilitation Assistant) [11] and KiReS (Kinect Rehabilitation System) [22] are exemplars of such tools, used for support of rehabilitation exercises. These systems employ a Kinect sensor to track patient movements, where a user interface displays two avatars that perform the prescribed exercise by the clinician and the ongoing movements and postures performed by the patient in real-time. Such visual feedback assists patients in improving their exercise performance, as well as in taking self-corrective action when needed [23]. Furthermore, the recordings of the daily exercise sessions can be sent via the internet to the respective clinician, who can assess the performance and provide feedback or corrective recommendations.

Likewise, traditional clinical assessment of patient progress is often based on a clinician's visual observation of patient movement or exercise performance [24]. Commonly used tests for this purpose may be condition-specific (e.g., FMA [Fugl-Meyer assessment], WMFT [Wolf motor function test], the ratio of optimal motion execution [11], [22]), more general screening of movement competency or performance (e.g., FMS [Functional Movement Screen$^{TM}$]), an evaluation of a specific muscle or joint (e.g., manual muscle test, range of motion testing), or a generic evaluation of specific skill of sport performance [25]. Some commonly utilized clinical evaluations or screening tools may be more objective and quantitative in nature, while others may rely more on a clinician's intuitive understanding and subjective rating of the patient's performance. Clinical tests/evaluations that are more subjective may have issues with the reliability and validity of the evaluation depending on the movement, exercise, or task being assessed [25]–[27]. Due to the subjectivity and challenge of evaluating many of the exercises and movements, sensors and data analytics may support clinical assessment by providing a complementary objective and quantitative measure of the quality of patient performance. For example, Oña Simbaña et al. [28] reviewed the systems for automated assessment of upper limb motor function using standard clinical tests.

To address the challenges associated with *home-based* and *in-clinic rehabilitation programs*, the development of systems that can reliably capture human movements, automatically analyze the recorded data, and evaluate the quality of the movement performance is critical. The provision of low-cost sensors with integrated functionality for tracking human motions provides an opportunity for the development of such systems. Furthermore, devising efficient computational algorithms for modeling and analyzing human motions becomes central to solving the problem of rehabilitation evaluation.

With regard to the sensory perception aspect of movement evaluation, obtaining precise movement data by motion sensors is crucial. Although a standard vision camera can be used as a motion sensor [29], [30], these cameras provide only 2-dimensional information about the captured scene and the lack of the third dimension's information imposes limits on the evaluation accuracy. To cope with this deficiency, one alternative is to use optical motion tracking systems, which employ a set of markers attached to strategic locations on a patient's body that are tracked by multiple high-resolution cameras [31]. These systems rely on computational algorithms to reconstruct the 3-dimensional scene by comparing and aligning the images taken by the set of multiple cameras. Although optical trackers are highly accurate and reliable, the high cost and need for attaching a set of markers during every session render them unsuitable for most cases of rehabilitation evaluation. Recent technology for 3-dimensional scene reconstruction based on vison/depth cameras has become popular due to the low cost and ease of use. Among the commercial vision/depth sensors, Microsoft Kinect [18] has been the preferred choice in most related works. Inertial sensors and accelerometers have also been extensively used for motion tracking and evaluation [32], due to their low cost and simple principles of operation. Skeletal data extracted from color/depth cameras or inertial sensors are widely used in the domain of rehabilitation evaluation. Such data consist of time-ordered sequences of angular or position coordinates of the joints in the human body. Full-body skeletal data are highly redundant, thus, they are rarely applied directly for modeling and analysis of human motions. Consequently, *feature engineering* via selection of important skeletal dimensions or distances is often employed for extracting relevant information from skeletal data [33]–[35]. Another common avenue for feature engineering entails creating new local features for representing the motions, based on a set of kinematic parameters that are predefined for an exercise, or by employing a functional mapping [36], [37]. Similarly, an often used processing step in feature engineering is dimensionality reduction, where unimportant or highly correlated dimensions are excluded from the data [38]–[41]. Principal component analysis (PCA) and its variants are widely used for dimensionality reduction of movement data [42]. Recently, a line of work emerged that uses machine learning models for automated extraction of features from collected rehabilitation data [39], [43], [44]. The efficient feature engineering of these algorithms produces an improved representation of the raw input data for movement evaluation, and it is an important constituent of the pertinent methods.

In this review, we adopted the following taxonomy for the main categories of computational methods for evaluation of rehabilitation exercises using motion capture data: discrete movement score, rule-based and template-based approaches. *Discrete movement score approaches* [45]–[48] classify individual repetitions of rehabilitation exercises into discrete classes, e.g., correct or incorrect. Conventional machine learning classifiers are commonly employed for this task, where the outputs are discrete class values of 0 or 1 (i.e., incorrect or correct repetition). A shortcoming of these methods is the inability to detect subtle changes in patient performance and provide intermediate levels of movement quality (for instance, with scores between 0 and 1). *Rule-based approaches* [36], [49]–[51] utilize a set of rules for a considered rehabilitation exercise defined in advance by clinicians or human movement experts. The rules are used as a gold standard for evaluating the level of correctness. A disadvantage of these approaches is that the rules are exercise-specific, and cannot be reused for other exercises. *Template-based approaches* are based on comparisons of measured movements with a template of the movements. The template is typically obtained from correct performance of the exercises by healthy subjects. One group of respective approaches employs distance functions for calculating a similarity score between patient-performed repetitions and reference template repetitions: Euclidean, Mahalanobis, and dynamic temporal warping distances are the most frequently used functions for this purpose [52]–[59]. The benefit of the distance functions is that they are not exercise-specific, and thus can be applied for evaluation of new types of exercises. Another line of works exploits the ability by probabilistic models with latent variables to encode spatial variability and temporal dynamics of human movements [39]–[41], [54], [60]–[62]. For instance, Gaussian mixture models [39], [63] and hidden Markov models [61], [62] were used for motion modeling, and quality assessment is based on the likelihood that the individual sequences are being drawn from a derived model. Whereas the probabilistic models are advantageous in handling the variability due to the stochastic character of human movements, models with abilities for a hierarchical data representation (such as deep neural networks [64]) can produce more reliable outcomes for movement quality evaluation, and better generalize across individual patients and musculoskeletal conditions.

In the context of objective evaluation of patient movements in rehabilitation programs, it is important to note that a large number of robotic and mechanical devices have been designed and are commonly used for quantitative movement evaluation. Accordingly, multiple review papers have provided overviews of related works in the literature [15], [65]–[70]. The review of rehabilitation assessment using robotic and mechanical devices is beyond the scope of this work, and we place emphasis on quantitative evaluation using computational approaches for analysis of movement data collected with motion capture sensors.

This review is organized as follows. Motion sensors for capturing rehabilitation exercises are reviewed in Section 2. Section 3 presents the related benchmark datasets for rehabilitation movements. Sections 4 discusses feature engineering for movement analysis. Section 5 details the various approaches for evaluation of patient movements during the performance of rehabilitation exercises. Scoring functions used to scale and adjust movement performance quantities and movement segmentation are discussed in the ensuing two sections. Potential future research directions are presented in Section 8. The final section briefly summarizes and concludes the paper.

## 2. Motion Capture Sensors

In general, a motion sensor is a device, module, or subsystem used to detect physical movement within an environment in real-time. Over the last decades, various advanced sensors for capturing human movement were developed. Currently, two types of motion sensors are widely used for rehabilitation exercise evaluation: optical and inertial sensors. Further, the respective optical sensors can be classified into two broad categories: vison/depth cameras and marker-based motion capture systems.

*2.1. Vision/depth Cameras*

There are two main types of depth cameras: structured light (SL) and time-of-flight (ToF) cameras. The commercial products Kinect v1 and Asus Xtion belong to the former. Kinect v2 and Azure Kinect are based on the ToF principle. SL cameras have lower cost, as they are simpler to construct. On the other hand, ToF cameras are more expensive, but are less affected by light variations, and therefore can be used in an outdoor environment.

*Kinect v1*—Microsoft launched Kinect for Xbox 360 as a gaming console in 2010, and the corresponding hardware version of the device for Windows (called Kinect v1) was released in 2012. The sensor includes a vision (RGB) camera, an SL depth camera, multiple microphones, and a motorized tilt. The SL camera employs an infrared laser projector combined with a monochrome Complementary Metal Oxide Semiconductor (CMOS) sensor to provide depth (i.e., range) information. Kinect v1 outputs synchronized 640×480 RGB and 320×240 depth images at a frame rate of 30 Hz. The combination of RGB and depth streams is often referred to as RGB-D data. Regarding the motion capture capabilities, Kinect v1 can provide 3D coordinates for 15 or 20 joints of a moving subject by using either OpenNI SDK (Software Development Kit) or Microsoft SDK, respectively. Due to its versatility and low price, the device was widely utilized for measuring human movements across different applications [20]–[22], [49], [52], [55], [56], [71]–[75].

Accordingly, many studies were conducted to verify the validity of Kinect v1 for posture measurement or motion capturing in biomedical applications [76]–[83]. The studies by Clark et al. [79], [83] were among the earliest works that evaluated the suitability of Kinect v1 for biomedical applications by comparing the measurements by Kinect v1 and a Vicon optical tracking system (used as a gold standard). Clark et al. [79] reported that the agreement between the Kinect v1 and Vicon ranged from excellent for parameters like gait speed, step length, and stride length (Pearson correlation > 0.9) to moderate for other parameters like the stride time (Pearson correlation = 0.69). In a similar study [83], the authors assessed the reliability of Kinect v1 for postural control in lateral reach, forward reach, and single-leg standing balance. The findings indicated that Kinect v1 exhibited excellent correlation for almost all measurements (Pearson correlation = 0.96; range, 0.84–0.99) and can validly be used for evaluating postural changes in a clinical setting. Galna et al. [80] studied the accuracy of Kinect v1 for measuring movement symptoms in people with Parkinson's disease and found that Kinect v1 measured the timing of the movements very accurately (Intraclass Correlation Coefficient (ICC) > 0.9 and Pearson correlation > 0.9), whereas for measuring the spatial characteristic of movements the agreement ranged from excellent (for gross movements such as sit-to-stand ICC = 0.989) to poor (for fine movements such as hand-clasping ICC = 0.012). Still, the authors reported high correlation between the spatial measurements for Kinect and Vicon for all movements (Pearson correlation > 0.8). Similarly, Tao et al. [78] estimated the root-mean-squared error (RMSE) between the measurements by Kinect v1 and an Optotrak optical tracker for hand reaching positions, trunk positions, and elbow angular orientations, and reported RMSE errors of 6.3 cm (2.5 inch), 9.8 cm (3.9 inch), and 26.7 degrees, respectively. Mishra et al. [81] designed a remote home-based rehabilitation program that uses Kinect v1 for motion tracking and streams the recorded videos in real-time to a clinic. The viability of the system was compared to a Vicon optical tracker. The authors introduced metrics for quantifying the angular trunk sway and reported a maximal measurement error of 17.2 degrees in anteroposterior (AP) direction and 7.3 degrees in mediolateral (ML) direction. In summary, almost all studies for validation of Kinect v1 reported an excellent temporal accuracy, whereas the spatial accuracy was high for larger body movements, and moderate to poor for more delicate movements. Other important considerations in practical applications of the sensor include: the accuracy of the measurements is dependent on the distance from the sensor and the selected view point for a particular movement, and limb occlusions during movements may impact the measurement accuracy of other body parts.

*Kinect v2*—Microsoft Kinect for Xbox One, an upgraded version of Kinect v1, was released in 2013. The corresponding hardware for Windows with the supporting SDK was released in 2014 (known as Kinect v2). Kinect v2 has a similar construction to Kinect v1, except that the SL depth sensor is replaced with a ToF depth sensor. The ToF sensor obtains the depth information by measuring the time it takes for an infrared laser pulse to travel back and forth between the camera and the surrounding objects. Kinect v2 offers an improved resolution of 1920×1080 RGB and 512×424 depth images at a frame rate of 30 Hz. The open-source Microsoft SDK2 allows extracting the skeletal coordinates of 25 joints. Kinect v2 has other advantages over its predecessor; for instance, it can detect objects up to 3 feet from the sensor, compared to 6 feet for Kinect v1.

Similarly, the precision of Kinect v2 for motion tracking has been assessed and reported in numerous publications [77], [81], [84]–[89]. For example, Napoli et al. [89] reported that the sensor can provide accurate joint displacements for a range of clinical tasks with RMSE of 3.4 cm (1.3 inch), while lower accuracy was noted when capturing joint angles with RMSE of 24.6 degrees. Comparable validation results were reported by Capecci et al. [34], where for a set of three rehabilitation exercises an average positional RMSE of 3.3 cm (1.3 inch) and an average orientational RMSE of 12.7 degrees were recorded. In [88], Otte et al. assessed the validity of Kinect v2 for clinical motion analysis by comparing its accuracy against a Vicon system, and concluded that the accuracy of measurements is moderate to excellent, where for most clinical parameters there was excellent consistency between the two systems. Dolatabadi et al. [85] employed two-group mean differences, i.e., Bland-Altman Limit of Agreement (LoA), and Intraclass Correlation Coefficient (ICC) between Kinect v2 and a GAITRite mat to verify the capacity of Kinect v2 for recording gait. In three walking conditions, for all gait parameters the maximum values of the group mean differences were 8%, the 95% LoA was less than or equal to 11%, and the ICC ranged between 0.9 and 0.98. The comparison results implied that Kinect v2 is capable of measuring spatio-temporal gait parameters for objective evaluation. In [81], trunk sway measures calculated from 3D joint positions were employed for validation. The performance across all trials had the maximum trunk sway error of 12.8 degrees in AP direction and 5.6 degrees in ML direction. Conclusively, Kinect v2 provided greater accuracy than Kinect v1, and most studies reported low measurement errors and adequate motion tracking abilities for a number of biomedical applications.

*Azure Kinect*—The latest generation of Kinect, called Azure Kinect, was released in July 2019. As the name suggests, the sensor's functionality is based on the integration with Microsoft's cloud computing service Azure. The target audience for this generation of the sensor is developers and businesses interested in artificial intelligence applications. The sensor offers an RGB camera with 3,840×2,160 pixels, a ToF depth camera with 1,024×1,024 pixels, an inertial measurement unit, and seven microphones. The depth camera supports five modes of operation, and the RGB camera offers six modes of operation, each with different resolutions and frame rates. Compared to Kinect v2, Azure Kinect is much smaller and lighter, the resolution of RGB/depth cameras is doubled and they support different modes and color formats, and it also provides inertial data apart from the RGB and depth data. More importantly, Azure Kinect is no longer a game peripheral (e.g., for Xbox One) but a smart device used primarily by developers with Azure cloud.

*Other Vision/Depth Sensors*—Asus Xtion Pro and Asus Xtion Pro Live both provide an SL camera for capturing depth information. Asus Xtion Pro Live houses also a color RGB camera and two microphones. The sensors are supported by OpenNI SDK for tracking 15 body joints. The accuracy of depth data recorded by Asus Xtion is studied in [90], where under seven lighting conditions, the reported median error was less than 1.3 cm (0.5 inch). Other commercial vision/depth sensors include Intel's RealSense [91] and Structure Sensor [92]. However, they are rarely used exclusively for motion tracking, and almost no efforts have been made to verify their accuracy and reliability for this purpose.

An overview of the main characteristics of the above vision/depth sensors used for motion capturing is provided in Table 1.

*2.2. Optical Motion Tracking Systems*

Vicon, OptiTrack [93], Optotrak, and PhaseSpace motion capture [94] are the most common optical motion capture systems. They employ a set of markers that are attached to predefined locations on the human body, while multiple cameras positioned at different viewing angles track the markers' locations during the movements. A dedicated software program utilizes trigonometrical relations among the markers in captured images and the locations of the cameras to calculate the positions and orientations of the body joints. Many studies confirmed the excellent positioning performance of these motion capture systems in both static and dynamic tests [95]–[97]. As a result, optical tracking systems are regarded as the gold standard for verifying tracking reliability of other motion sensors [76], [98]–[100]. On the other hand, their high cost limits their broad applicability for the assessment of patient rehabilitation progress.

*2.3. Inertial Sensors*

A general form of an inertial sensor includes an accelerometer and a gyroscope [101]. Integrated devices containing inertial sensors are collectively called inertial measurement units (IMUs). It should be noted that inertial sensors described in this review only refer to wearable sensors, and inertial data means measurements provided by accelerometers or inertial sensors. The accelerometer is a compact device designed to measure non-gravitational acceleration, which is the rate of change of the sensor's positional velocity. The sensor's position is typically obtained by first subtracting the earth's gravity from accelerometer measurements and afterward applying double integration. A gyroscope records the angular velocity of the sensor, i.e., the rate of change of the sensor's orientation [102]. The measured positions and orientations of wearable inertial sensors are often used for analysis of human postures and motions. For the purpose of this study, joint positions and angles transformed from inertial sensors are still considered skeleton data, despite its sparse representation of the human motions.

Analogously to the previously described sensors, prior research focused on verifying the validity of inertial sensors for rehabilitation analysis and evaluation. Chung and Ng [103] concluded that accelerometers measure motor reaction times with a large ICC value of 0.74 ($p < 0.001$). Yet, the reaction time measured by the accelerometer is on average 8 milliseconds slower than that detected by the Vicon system. The work by Lugade et al. [104] reports that median sensitivities of an algorithm benchmarked on data provided by tri-axial accelerometers for activity identification are over 85% accurate in comparison to human raters. Fortune et al. [105] studied the ability of accelerometer data to be used for step counting. During walking or jogging tests, activity monitoring achieved a high median agreement of 92% with an interquartile range of 8%, which outperformed FitBits and a Nike Fuelband. The publications [106] and [107] both focus on the validity and reliability of inertial sensors for recording trunk movements. The former used the Pearson correlation and RMSE as metrics to measure the agreement between inertial and Optotrak measurements. The median values of the Pearson correlation exceeded 0.95, and the RMSE had a median value of 1 and 1.2 degrees in the AP and ML directions, respectively. In the latter, the coefficient of determination ($r^2$) and RMSE were adopted to validate the IMU system. In the primary movement direction, the RMSE ranged between 1.1 and 6.8 degrees, and the coefficients of determination were no less than 0.85. Washabaugh et al. [108] used Lin's concordance correlation coefficient (i.e., LCC [101]) and the Pearson correlation coefficient to evaluate IMU's validity for capturing spatiotemporal gait metrics. The authors concluded that inertial sensors provided accurate measurements for the respective study, however the degree of accuracy and reliability relied on several factors, such as the sensor position and movement speed. The validation studies in the literature generally agree that inertial sensors provide sufficiently accurate and fast movement data for rehabilitation analysis and evaluation.

*2.4. Other Sensors*

Other devices and sensors have been used in several works for rehabilitation evaluation. For instance, although standard vision cameras do not provide sufficient accuracy for motion capture, they can be employed to acquire facial expressions for pain detection [109], [110]. In [44], sequences of pressure maps were generated by a pressure-sensitive bedsheet to identify bed rehabilitation exercise. Teague et al. [111] performed joint health evaluation by sensing acoustical emissions from the knee using three types of microphones. And, in [112], thermal infrared images of patients were recorded by a medical infrared camera system and a quantitative evaluation of pain-related thermal dysfunction was obtained by analyzing the distribution of the skin temperature.

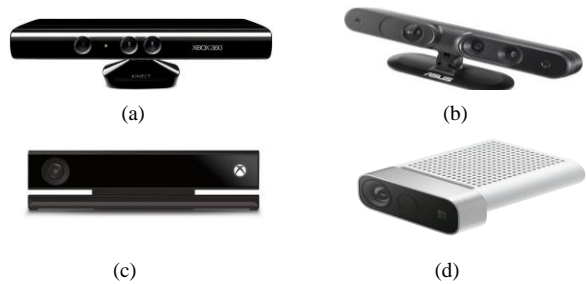

Figure 1. (a) Kinect v1; (b) Asus Xtion PRO LIVE; (c) Kinect v2; (d) Azure Kinect DK.

Table 1. RGB-D sensor capability comparison.

| **Feature** | **Kinect v1** | **Asus Xtion PRO LIVE** | **Kinect v2** | **Azure Kinect DK** |
|---|---|---|---|---|
| RGB camera | 1,280×960 px at 12 Hz<br>640×480 px at 30 Hz | 1,280×1,024 px at 30 Hz | 1,920×1,080 px at 30 Hz | 3,840×2,160 px at 30 Hz |
| Depth camera | 320×240 px at 30 Hz | 640×480 px at 30 Hz<br>320×240 px at 60 Hz | 512×424 px at 30 Hz | 640×576 px at 30 Hz<br>512×512 px at 30 Hz<br>1,024×1,024 px at 15 Hz |
|  | SL | SL | ToF | ToF |
| Motion sensor | None | None | 3-axis accelerometer | 3-axis accelerometer + 3-axis gyroscope |
| Measuring range | 0.85~4m | 0.8m~3.5m | 0.5~4.5m | 0.5~3.86m; 0.5~5.46m;<br>0.25~2.88m; 0.25~2.21m |
| Field of view | 57×43 degrees | 58×45 degrees | 70×60 degrees | 75×65 degrees;<br>120×120 degrees |
| Skeleton joints | 15 or 20 joints | 15 joints | 25 joints | 32 joints |

## 3. Datasets

A large number of publicly available datasets related to general human movements collected with healthy subjects are available for analysis [113]–[116]. The datasets are extensively used for benchmarking algorithms for action recognition, gesture recognition, or pose estimation. On the other hand, collecting large data sets of rehabilitation exercise data from patients suffering from an impairment or injury is more challenging due to privacy and safety concerns. Consequently, only a few public datasets for rehabilitation evaluation currently exist, and their main attributes are summarized in Table 2. The table lists the referenced publications, used sensors, data modality, number of subjects performing the exercises, number of exercises, and data availability.

*Taylor et al.*—For the dataset collected by Taylor et al. [45], experimental data is obtained from 9 subjects performing 3 exercises: standing hamstring curl, reverse hip abduction, and lying straight leg raise. Five inertial sensors (accelerometers) are positioned on the thigh and shin of both legs and the waist to measure the 3-axis acceleration of the corresponding body part. Each exercise is repeated by each subject 10 times, both on the left and right sides, respectively. The acquired data is manually segmented into the individual repetitions of each exercise.

*PAMAP2*—Physical Activity Monitoring Dataset (PAMAP2) [117], [118] was designed for activity recognition and estimation of exercise intensity (which involves light, moderate, or vigorous intensity of effort). It consists of 3,850,505 instances, recorded by 3 inertial sensors and a heart rate (HR) monitor. The sampling frequency of the inertial sensors and HR monitor are 100 Hz and 9 Hz, respectively. The data collection involved 9 subjects (8 males and 1 female) performing 18 different physical activities. The duration of each activity is between 1 and 3 minutes.

*SPHERE-Staircase2014*—This dataset [119] consists of 48 video sequences captured by an Asus Xtion camera, whereas the associated skeleton data is obtained with OpenNI SDK. The activities include walking upstairs in normal or abnormal gaits. There is a total of 12 persons completing this process. The abnormal gaits include freezing of gait and using a leading leg (left or right leg). A qualified clinician manually labeled each frame as normal or abnormal performance. The data is provided in the form of skeleton time-series.

*HPTE dataset*—Home-based Physical Therapy Exercises (HPTE) dataset by Ar and Akgul [71] contains 240 color/ depth video streams captured with Kinect v1. Although the RGB and depth videos collected by the Kinect sensor are 640×480 pixels, they are stored as 256 gray-level images in 320×240 pixels size. Five volunteers were tasked to perform eight exercises, where each subject performed an exercise 6 times consecutively, resulting in 30 repetitions per exercise. The duration of each repetition is between 15 and 30 seconds.

*dataELEMENT*— Created by Cuellar et al. [72] and collected with Kinect v1, dataELEMENT includes movements by 10 healthy subjects performing 5 exercises, where each exercise was repeated 10 times. The recorded data comprises absolute angles of joints or bones with respect to an underlying 3D base coordinate system, and relative angles between the bones that share a joint.

*Kinect 3D Active*—The dataset was created by Leightley et al. [120], [121] and it contains over 225,000 frames of depth and skeleton data recorded with Kinect v2 in a lab-based indoor environment. The dataset includes the subjects' personal information and the tracking states of all joints (i.e., 'tracked', 'not tracked', and 'inferred' states). Fifty-four participants (32 men and 22 women) participated in the data collection, and they were required to take standardized tests, including Short Physical Performance Battery (SPPB) [122], Timed-Up-and-Go (TUG) [123], vertical jump, and balance tests.

*UI-PRMD*—University of Idaho – Physical Rehabilitation Movement Dataset (UI-PRMD) created by Vakanski et al. [124] consists of 10 exercises that are widely applied in physical rehabilitation programs: deep squat, hurdle step, inline lunge, side lunge, sit to stand, standing active straight leg raise, standing shoulder abduction, standing shoulder extension, standing shoulder internal-external rotation, and standing shoulder scaption. Ten healthy subjects performed each exercise 10 times in a correct and incorrect manner. Two sensors were employed for collecting the data: a Vicon optical tracking system and Kinect v2. The Vicon system provided positions and orientation angles for 39 joints, whereas Kinect measured the positions and orientation angles for 22 joints of the exercise movements.

*KIMORE*—KInect-based MOvement Rehabilitation dataset (or KIMORE) by Capecci et al. [125] was collected with Kinect v2, and involves 78 subjects performing 5 exercises that are clinically recognized for low back pain physiotherapy. The exercises are: lifting the arms, lateral tilt of the trunk with arms in extension, trunk rotation, pelvis rotations on the transverse plane, and squatting. The enrolled population consisted of 44 healthy subjects and 34 patients with chronic motor disabilities. For each exercise, clinicians defined rules for extracting features from the raw data, with the corresponding features provided in the dataset. Additionally, the clinical scores derived from a clinical questionnaire [126] for evaluating the subjects' movement performance are included in the dataset.

Table 2. Datasets description.

| Dataset | Ref. | Year | Sensor | Modality | Subjects | Exercises | Available |
|---|---|---|---|---|---|---|---|
| Taylor et al. | [45] | 2010 | 5 inertial sensors | Inertial data | 9 | 3 | No |
| PAMAP2 | [118] | 2012 | 3 inertial sensors, heart rate monitor | Inertial data; heart rate signal | 9 | 18 | Yes |
| SPHERE-Staircase2014 | [119] | 2014 | Asus Xtion | Depth video; skeleton data | 12 | 18 | Yes |
| HPTE | [71] | 2014 | Kinect v1 | Gray-level and depth video | 5 | 8 | Yes |
| dataELEMENT | [72] | 2014 | Kinect v1 | Skeleton data | 10 | 5 | Yes |
| Kinect 3D Active | [121] | 2015 | Kinect v2 | Depth and skeleton data | 54 | 13 | Yes |
| UI-PRMD | [124] | 2018 | Kinect v2 + Vicon | Skeleton data | 10 | 10 | Yes |
| KIMORE | [125] | 2019 | Kinect v2 | RGB and depth videos; skeleton data | 78 | 5 | Yes |

## 4. Feature Engineering

Feature engineering is the process of creating features from raw data in order to improve the performance of a computational method. In rehabilitation evaluation, the recorded data from motion capture sensors are often in the form of high-dimensional time-series sequences of the joints' locations and/or orientations. Such data is exceedingly redundant and correlated (e.g., the wrist displacements are highly correlated to the elbow displacements), and they are rarely applied directly for modeling and analysis of rehabilitation movements. Accordingly, feature engineering via selection of important joints or limb distances to extract lower-dimensional representations from the raw data is often applied as a data processing step in rehabilitation evaluation.

In general, *feature engineering* involves feature extraction and feature selection. *Feature extraction* refers to generating new features from raw input data by a functional mapping. The mapping can be determined manually by experts, or it can be learned from existing data representations. *Feature selection* entails selecting the most important features among existing and/or extracted features. The set of newly selected features is a subset of the original features. Efficient feature engineering produces an improved representation of the input data for the underlying task.

In many related works, feature engineering is performed manually based on authors' understanding of human movements [33]–[37], [54], [97], [98], [126], [127]. For example, in [34], underarm angles and Euclidean distance between the elbows were used to describe the lifting of the arms. Similar, for squatting evaluation, knee angles and Euclidean distance between the ankles were extracted as clinical features. Jung et al. [37] proposed a set of distinctive features obtained from experimental data, consisting of mean speed, reaction time, duration, peak velocity, maximum velocity, distance error, direction error, and path length ratio. For assessing upper body movement after stroke, the range of motion, movement speed, symmetry ratio among body sides, and vertical distance were adopted as movement performance indicators in [35]. Yu and Xiong [128] selected eight bone vectors as important features to an algorithm for producing quality scores in support of home-based rehabilitation.

Although the approaches based on manual feature selection benefit from the authors' intuitive understanding of the most important attributes for particular motions, creating features manually requires domain-knowledge and is time-consuming.

*Automated feature engineering* to some degree can obviate the need for manual feature engineering. PCA is one of the most popular approaches for this task. It uses an orthogonal transformation to project measured correlated variables into linearly uncorrelated variables. Jun et al. [47] applied PCA on raw motion data consisting of joint positions for dimensionality reduction. However, PCA is not suited to identify the nonlinear structure of data [129], and therefore feature engineering based on nonlinear mapping has been studied extensively. Huang et al. [44] employed two types of nonlinear manifold learning—Local Linear Embedding and Isomap— for reducing the dimensionality of images. Another manifold-based approach using diffusion maps was employed for dimensionality reduction of skeletal data by Paiement et al. [40]. This method can find meaningful geometric descriptions of the data and has robustness to noise (Coifman and Lafon [38]). The nonlinear transformation of inputs by autoencoder neural networks has been used for dimensionality reduction of skeletal data [39]. Similarly, Crabbe et al. [43] proposed a CNN-based algorithm to extract a low-dimensional pose representation from depth images.

In many prior works, feature engineering has been realized in two or three consecutive steps. Researchers typically first define new features from the raw data and then apply automated feature engineering to simplify these features. For instance, several time-domain and frequency-domain features were extracted from inertial data in [130], and afterward PCA was applied to reduce the overall number of features for training. Houmanfar et al. [53] first derived statistical features from motion sequences, and after that, Least Absolute Shrinkage and Selection Operator (LASSO) was utilized to select five most relevant features. In [41], Tao et al. first constructed four possible feature descriptors by the geometry of the human body, and afterward, extracted the final features for each descriptor by using the method described in [40].

Manual feature selection is very common in movement data analysis and evaluation, since human understanding of the importance of specific features of the rehabilitation exercises provides excellent leverage toward the design and initialization of computational methods. Nevertheless, human movements are very complex, resulting in high-dimensional data with intricate temporal and spatial dependencies between the joint positions and orientations, and involve numerous constraints between the joints which are difficult to accurately encode or define manually. We hold that the approaches for automated feature engineering offer more powerful means for learning the underlying correlations and constraints in movement data than the approaches for manual feature selection. In this respect, deep learning models are the most compelling approaches for feature engineering at present, because of the ability to automatically learn spatio-temporal features at multiple levels of abstraction in high-dimensional data.

The reader is referred to Table 3 for a complete list of publications based on the sensors, data representation modalities, selected features, and pertinent tasks.

## 5. Evaluation Methods

Successful motion quality evaluation in rehabilitation programs depends on efficient quantification of the level of performance of rehabilitation exercises from measured motion data. The approaches for evaluating rehabilitation exercises can generally be categorized into: discrete movement score approaches, rule-based approaches, and template-based approaches. A summary of the approaches, advantages/disadvantages, and referenced works for each category is provided in Table 4.

*5.1. Discrete Movement Score Approaches*

This category of approaches classifies individual exercise repetitions into several discrete classes. Most often, the discrete classes are: correct and incorrect movements. Thus, the outputs are typically binary class values of 0 or 1 (i.e., incorrect vs correct repetition). Adaboost classifier [45], $k$-nearest neighbors [48], Bayesian classifier [71], and an ensemble of multi-layer perceptron neural networks [37] have been used to distinguish between the two classes. For instance, in [47] $k$-nearest neighbors classifier was applied for exercise classification after filtering the data noise and applying dimensionality reduction through PCA. The approach achieved 95.6% classification accuracy. Similarly, machine learning classification was applied in prior research for movement classification into score categories for standards clinical tests [28]; e.g., Support Vector Machines (SVM) [131], random forest [132], and artificial neural networks [133] were used for automated FMA, naïve Bayes classifier [134] was implemented for WMFT, and random forest [135] was used for Functional Ability Scale (FAS) evaluation. Furthermore, Um et al. [46] utilized an ensemble of convolutional neural networks to detect Parkinson's disease states in data collected with a wrist-worn wearable sensor, where the states were defined as: OFF state with Parkinson's syndrome symptoms, DYS state with dyskinetic symptoms, and ON state with no salient Parkinson's syndrome or dyskinetic symptoms observed. The studies employing discrete movement scores have reported high accuracy in distinguishing correct from incorrect movement sequences. In spite of that, a shared shortcoming of the approaches is the lack of ability to monitor continuous changes in movement quality, or quantify the progress of patient performance over the duration of the rehabilitation program. Subsequently, discrete movement score category is less relevant to the development of systems for quantifying the rehabilitation performance.

*5.2. Rule-Based Approaches*

The class of rule-based approaches utilizes a set of rules for a considered rehabilitation exercise that is defined in advance by clinicians or human movement experts. The rules are used as a gold standard for assessing the level of correctness of the movements. Whereas a smaller number of rules, such as relative angles or distances, may be sufficient for representation of simpler movements, a more comprehensive set of rules is needed to describe more complicated exercises. For instance, the quality of sit-to-stand and squat exercises was measured by the knees and ankles angles in [50]. Similarly, in [49], three types of kinematic rules were defined to model rehabilitation exercises: rules for dynamic movement, rules for static postures, and rules for movement invariance. Afterward, fuzzy logic was applied to generate a single final score that represents the quality of the rehabilitation exercise. Exploring rule-based approaches for rehabilitation exercises is a valuable option for simpler exercises, however, it becomes increasingly more difficult to extract reliable features and obtain an objective evaluation for more complex types of rehabilitation exercises. In addition, these approaches lack flexibility and capacity for generalization to new exercises, since a different set of rules is required to be selected for each individual exercise.

*5.3. Template-Based Approaches*

In template-based approaches, patients' exercise performance is evaluated based on the difference between training motion sequences executed by the patients and template motion sequences. For example, the training sequences may be captured during a patient's practice, and the template sequences can be reference movements performed either by healthy subjects, clinicians, or by patients under a clinician's supervision. The metrics used to measure motion similarity in template-based approaches can be classified into two categories: distance functions, and probability density functions.

*a) Distance Functions*

Several distance function-based approaches have been used for movement evaluation. In [52], Euclidean distance between reference template positions and velocities and the user's positions and velocities were employed for calculating the motion similarity. Such an approach provided immediate feedback to the users performing the exercises via the position and velocity errors. Houmanfar et al. [53] measured the level of correctness of rehabilitation exercises performed by patients, by computing a Mahalanobis distance between the patient-performed repetitions and the mean value of a set of repetitions completed by a group of healthy subjects (used as the ground truth). Furthermore, the authors used the derived distance measure for the individual repetitions to develop additional metrics for quantifying the quality of a set of repetitions, a set of exercises, and for tracking patients' progress over a period of time consisting of several sessions. The progress of a group of 18 patients was monitored during a hospital stay ranging from 4 to 12 days, and it was found that the calculated progress scores correlated well with the physiotherapist's evaluation of the patients' performance.

To formulate distance functions, let's assume the notation $\mathbf{X} = \{x^{(1)}, x^{(2)}, \cdots, x^{(L)}\}$ and $\mathbf{Y} = \{y^{(1)}, y^{(2)}, \cdots, y^{(L)}\}$ for two motion sequences, where $x^{(t)}$ and $y^{(t)}$ are the joint measurement vectors at time $t$. The measurements at a particular time moment are multi-dimensional vectors denoted $x^{(t)} = (x_1^{(t)}, x_2^{(t)}, \cdots, x_D^{(t)})$ and $y^{(t)} = (y_1^{(t)}, y_2^{(t)}, \cdots, y_D^{(t)})$, where $D$ is the dimensionality of the data. *Euclidean distance* between two sequences $\mathbf{X}$ and $\mathbf{Y}$ is defined by $d_E(\mathbf{X}, \mathbf{Y}) = \sum_{t=1}^{L} \|x^{(t)} - y^{(t)}\| = \sum_{t=1}^{L} \sqrt{\sum_{d=1}^{D}(x_d^{(t)} - y_d^{(t)})^2}$. Similar, *Mahalanobis distance* between two sequences $\mathbf{X}$ and $\mathbf{Y}$ is $d_M(\mathbf{X}, \mathbf{Y}) = \sum_{t=1}^{L} \sqrt{(x^{(t)} - y^{(t)}) V^{-1} (x^{(t)} - y^{(t)})^T}$ where $V$ is the covariance matrix of the data. In fact, the Euclidean distance is a special case of the Mahalanobis distance when the covariance matrix is an identity matrix.

The main limitation of the Euclidean distance or its variants is the requirement for the compared motion sequences to have the same length. This drawback is overcome with the dynamic time warping (DTW) distance [136]. The DTW distance is the most commonly adopted for measuring motion dissimilarity between training sequences and reference template sequences [54]–[57], [128], [137]–[141]. For two univariate time series $x = (x_1, x_2, \cdots, x_m)$ and $y = (y_1, y_2, \cdots, y_n)$, a time warping path is a sequence $W = (w_1, w_2, \cdots, w_K), max(m, n) \leq K < m + n - 1$ where the element $w_k = (i, j)$ indicates the matching relationship between $x_i$ and $y_j$. The DTW distance between two sequences $\mathbf{X}$ and $\mathbf{Y}$ is $d_{DTW}(x, y) = \max_W \{\sum_{k=1}^{K} \delta_k(i, j)\}$, where $\delta_k(i, j) = |x_i - y_j|$ or $(x_i - y_j)^2$.

The DTW distance has been applied to a broad range of features extracted in movement data. For example, Saraee et al. [137] used the DTW distance and speed ratio of key body joints for an exercise to directly evaluate the quality of performance with respect to reference template sequences. In [138] and [123], movement quality scores were derived by scaling the DTW distance values in the [0, 1] range. In the former, the DTW distance errors were scaled by the lower bound and upper bound, based on a scoring function introduced in [142]. The latter used a sigmoid function to map the DTW distance error into the required range. The publication [55] used a Euclidean norm of the DTW difference for evaluating exercise movements. In [57], the DTW distance was calculated for a set of selected features, and afterward, an adaptive neuro-fuzzy algorithm was used for calculating the overall quality of the practice sequences. Likewise, the study by Yurtman and Barshan [139] introduced a multi-template multi-match DTW algorithm to measure the similarity between training sequences and previously recorded template sequences.

Other distance functions have also been reported in the literature. For instance, a distance function similar to the Hausdorff distance was proposed to measure the similarity between two exercise sequences [44]. Concretely, the distance between two sequences $\mathbf{X}$ and $\mathbf{Y}$ was defined as $d(\mathbf{X}, \mathbf{Y}) = s(\mathbf{X}, \mathbf{Y}) + s(\mathbf{Y}, \mathbf{X})$, where $s(\mathbf{X}, \mathbf{Y}) = \frac{1}{L}\sum_{i=1}^{L} \min_{1 \leq j \leq L} \|x^{(i)} - y^{(j)}\|$. Similarly, the coefficient of cross-correlation [143] between two skeleton sequences was used to measure motion similarity; for two motion sequences $\mathbf{X}$ and $\mathbf{Y}$ consisting of $N$ data points, the coefficient of cross correlation [144] is defined by: $r_{xy}(k) = \frac{c_{xy}(k)}{\sqrt{c_{xx}(0)c_{yy}(0)}}$, where $c_{xx}(0) =$ $\sum_{i=1}^{N}(x_i - \bar{x})^2$, $c_{yy}(0) = \sum_{i=1}^{N}(y_i - \bar{y})^2$, $c_{xy}(k) = \begin{cases} \sum_{i=1}^{N-k}(x_i - \bar{x})(y_{i+k} - \bar{y}) + \\ \sum_{i=N-k+i}^{N}(x_i - \bar{x})(y_{i-N+k} - \bar{y}), k = 1,2,\cdots,N \\ \sum_{i=1}^{N}(x_i - \bar{x})(y_i - \bar{y}), \ k = 0 \end{cases}$, and $k$ is an index indicating a time shift of one sequence with respect to the other. Another distance metric used for this purpose is the deep metric [145], which, differently from the Euclidean distance and DTW, can capture contextual information and semantic relationship between two motion sequences. The deep metric is defined as the Euclidean distance in an embedding space $f$, i.e., the distance between two motion sequences **X** and **Y** is $d_L(X,Y) = \|f(X) - f(Y)\|$.

The studies in the literature based on evaluation using distance functions reported a high correlation between patient's performance and clinicians' evaluation. For instance, DTW-based evaluation achieved high posture monitoring accuracy (91.9%) and exercise monitoring accuracy (95.2%) in comparison to clinicians' annotated rehabilitation data [55], high correlation with the Brunnstrom stages of recovery (86% at $p<0.001$) [138], and high predictive score accuracy (80%) [57] in comparison to clinical evaluation. The DTW distance is especially suitable for rehabilitation evaluation, since it can compensate for the variability and time-shift in movement sequences. Conclusively, the main advantage of the distance function approaches is that they are not exercise-specific, and hence can be applied for evaluation of new types of exercises. However, the distance functions also have shortcomings, because they do not attempt to derive a model of the rehabilitation data, and the distances are calculated at the level of the individual time-steps in the raw measurements.

*b) Probability Density Functions*

A body of research work utilized probability density functions to model and evaluate rehabilitation exercises, due to the abilities of probabilistic models for handling the stochastic variability of human movement. For instance, the log-likelihood of individual sequences drawn from a trained Gaussian mixture model has been used for movement quality evaluation [146], [146]. Discrete hidden Markov models (HMM) were implemented for analysis and segmentation of human motion data for rehabilitation exercises [53], [147]. In [148], an approach based on hidden semi-Markov models (HSMM) was applied to evaluate five different rehabilitation exercises and provide an evaluation score [61], [62]. The requirement for segmenting the exercises into individual repetitions by discrete HMM or HSMM was overcome in [40], [41] by employing a continuous HMM.

*GMM Log-likelihood*: A Gaussian mixture model (GMM) with $K$ Gaussian densities has the form $P(x) = \sum_{k=1}^{K} \pi_k \varphi(x|\mu_k, \Sigma_k)$, where $\varphi(x|\mu_k, \Sigma_k)$ is the $k^{\text{th}}$ Gaussian function with mean $\mu_k$ and covariance matrix $\Sigma_k$, and $\pi_k$ denotes mixing coefficients satisfying the constraint $\sum_{k=1}^{K} \pi_k = 1$. GMM is trained by maximizing the log-likelihood of the template sequences (e.g., which can be collected from a group of healthy subjects), which for a sequence **Y** is defined by $L(Y) = \log(\prod_{t=1}^{L} P(y^{(t)})) = \sum_{t=1}^{T} \log\{\sum_{k=1}^{K} \pi_k \varphi(y^{(t)}|\mu_k, \Sigma_k)\}$. The deviation between the motion sequences **X** and **Y** is $d_G(X,Y) = \frac{1}{L}|L(X) - L(Y)|$.

*HMM Log-likelihood*: Hidden Markov model (HMM) has $M$ possible states denoted $S = \{s_1, s_2, \cdots, s_M\}$ where the state at time $t$ is $q_t \in S$. The likelihood of a sequence **Y** is calculated as $P(Y) = \sum_{q_1, q_2, \cdots, q_L \in S} \pi_{q_1} P(y^{(1)}|q_1) \prod_{t=2}^{L} P(y^{(t)}|q_t) P(q_t|q_{t-1})$, where $\pi_{q_1}$ is the initial state distribution, $P(y^{(t)}|q_t)$ is the probability that the observation $y^{(t)}$ is seen if in state $q_t$, and $P(q_t|q_{t-1})$ is the transition probability from state $q_{t-1}$ to state $q_t$. The deviation between the motion sequences **X** and **Y** is calculated by $d_H(X,Y) = \frac{1}{L}\left|\log\frac{P(X)}{P(Y)}\right|$.

Employing a probability density function approach, Capecci et al. [62] produced one of the most complete recent works on rehabilitation evaluation, where the authors asked two clinicians to score a set of movements, and used the scores as a gold reference standard for validating an HSMM-based approach. The research reported a high correlation between the HSMM-generated and the clinicians' movement scores (Pearson correlation coefficient of 0.62, $p < 0.01$). Furthermore, in comparison to a DTW-based evaluation (Pearson correlation coefficient of 0.56, $p < 0.01$), the probabilistic HSMM model demonstrated better correlation with the clinicians' scores. In this work, a Kinect v2 sensor was used for motion capturing, and the authors indicated lower tracking errors for the upper body in comparison to the lower body. Similar results have been reported by other researchers, noting high correlation between machine learning-based and physiotherapist-based evaluation [53].

Utilizing probabilistic approaches for exercise evaluation is advantageous in comparison to all other evaluation methods, because they employ statistical probability distributions to handle the random variability of human movements. The ability to model the stochastic variations in performing the same exercise both by the same subject and across different subjects is essential for efficient movement modeling and evaluation. One shortcoming of the probabilistic models is that the movements are represented at a single level of movement abstraction, and it is difficult to implement probabilistic modeling at multiple levels of movement abstraction.

Table 3. Summary of sensors, modalities, features, and objectives per publication. G: gray-level image; D: depth image; S: skeleton data; I: inertial data; MC: motion classification; MA: motion assessment; MS: motion segmentation; PE: pose estimation; DA: data augmentation.

| Reference | Sensor | Modality | Feature engineering | Objective |
|---|---|---|---|---|
| Ar and Akgul [20] | Kinect v1 | G + D | Histogramming 3D Haar-like features | MC |
| Ar and Akgul [71] | Kinect v1 | G + D | Histogramming 3D Haar-like features | MC |
| Benetazzo et al. [52] | Kinect v1 | S | None | MA |
| Cuellar et al. [72] | Kinect v1 | S | Hand-crafted | MA |
| Hagler et al. [73] | Kinect v1 | S | Hand-crafted | MA |
| Nomm and Buhhalko [74] | Kinect v1 | S | Hand-crafted | MC |
| Zhao et al. [49] | Kinect v1 | S | Hand-crafted | MA |
| Antón et al. [22] | Kinect v1 | S | Hand-crafted | MA |
| Antón et al. [55] | Kinect v1 | S | Hand-crafted | MC |
| Su [56] | Kinect v1 | S | Hand-crafted | MA |
| Su et al. [57] | Kinect v1 | S | Hand-crafted | MA |
| Crabbe et al. [43] (data from [119]) | Kinect v1 | D | None | PE |
| Uttarwar and Mishra [75] | Kinect v1 | S | Hand-crafted | MC |
| Saraee et al. [137] | Kinect v2 | S | Hand-crafted | MA |
| Vakanski et al. [39] (data from [113]) | Kinect v2 | S | Autoencoder network | MA |
| Paiement et al. [40] (data from [182]) | Kinect v2 | S | Diffusion maps | MC + MA |
| Parisi et al. [182] | Kinect v2 | S | Hand-crafted | MA |
| Capecci et al. [33] | Kinect v2 | S | Hand-crafted | MA |
| Capecci et al. [34] | Kinect v2 | S | Hand-crafted | MA |
| Capecci et al. [126] | Kinect v2 | S | Hand-crafted | MA |
| Capecci et al. [61] | Kinect v2 | S | Hand-crafted | MA |
| Capecci et al. [62] | Kinect v2 | S | Hand-crafted | MA |
| Tao et al. [41] (data from [119]) | Kinect v2 | S | Diffusion maps | MC + MA |
| Osgouei et al. [54] | Kinect v2 | S | Hand-crafted | MC |
| Spasojević et al. [35] | Kinect v2 | S | Hand-crafted | MA |
| Yu and Xiong [128] | Kinect v2 | S | Hand-crafted | MA |
| Saraee et al. [156] | Kinect v2 | S | Hand-crafted | MA |
| Taylor et al. [45] | Inertial sensor | I | Hand-crafted | MC |
| Zhang et al. [48] | Inertial sensor | I | Cross-correlation function | MC |
| Lin and Kulić [148] | Inertial sensor | S | None | MS |

| Chen et al. [183] | Inertial sensor | I | Hand-crafted | MC |
| Zhang et al. [138] | Inertial sensor | I | None | MA |
| Houmanfar et al. [53] | Inertial sensor | I | Hand-crafted + LASSO | MA |
| Msayib et al. [184] | Inertial sensor | S | Hand-crafted | MA |
| Um et al. [185] | Inertial sensor | I | None | DA |
| Um et al. [186] | Inertial sensor | I | None | MC |
| Um et al. [46] | Inertial sensor | I | None | MC |
| Burns et al. [187] | Inertial sensor | I | HAR statistical and heuristic features [188] | MC |
| Yurtman and Barshan [139] | Inertial sensor | I | None | MC + MA |
| Karime et al. [127] | Inertial sensor | I | Hand-crafted | MA |
| Coskun et al. [189] | Vicon, ToF + camera | S | None | MC |
| Vamsikrishna et al. [190] | Leap motion controller | S | Hand-crafted | MC |

Table 4. Summary of approaches for evaluation of rehabilitation movements.

| Approach | Advantage | Disadvantage | Reference |
| --- | --- | --- | --- |
| Discrete movement score approaches | Efficient and achieve high accuracy | Cannot track diverse degrees of functional abilities | [37], [45]–[48], [55], [71], [74], [106], [183], [185]–[187] |
| Rule-based approaches | Provide multiple performance scores; less computation complexity | New rehabilitation exercises require different rules to be designed | [33]–[36], [49], [50], [61], [62], [73], [126], [127], [156], [182], [191]–[195] |
| Template-based approaches | Avoid the process of making rules; Can reflect the level of motor ability | Only give a overall score for exercise performance | [39]–[41], [52], [53], [55], [57], [96], [97], [103], [128], [138], [140], [143], [145], [148], [156], [190], [196] |

## 6. Movement Segmentation

The objective of exercise segmentation is to extract individual repetitions from a continuous motion sequence of an exercise. Movement segmentation is an important step for evaluation of physical rehabilitation exercises, because most of the existing evaluation approaches are based on quantifying the quality of individual repetitions of an exercise. Consequently, after a patient's movements are recorded with a motion capture sensor (and the patient performed multiple repetitions of the exercise), it is first required to segment the motion data into the instances of the individual repetitions, and only afterward is the evaluation technique applied to produce a quality score for the individual repetitions. The overall quality of the exercise is usually calculated by averaging over the performance scores of the individual repetitions [148].

Although in many studies the motion sequences are segmented manually, such an approach is not conducive to the realization of fully automated evaluation of rehabilitation exercises. Existing approaches for automated motion segmentation are broadly classified into two categories: (1) approaches that model the common characteristics shared by segment points, and (2) approaches that learn a segment pattern from a template library. In the first class, kinematic zero crossing (KZC) methods are frequently used to perform exercise segmentation. These methods determine segments based on zero crossings for the velocity [61], [62] or acceleration [149] of joint trajectories. Distance functions, such as Euclidean distance [150], Mahalanobis distance [151], and DTW distance [152] have also been used for this purpose, where segments are extracted at the points having the value of the distance function greater than a pre-selected threshold. Lee et al. [153] introduced a deep learning-based approach for segmentation of time-series, in which an autoencoder network extracted representative features from input data, and the peaks in a distance function calculated from the features were selected as breakpoints for segmentation purpose. These methods rely on domain-specific knowledge of the underlying data to select discriminative features for segmentation purpose, and do not offer a mechanism to reject false positives. Thus, further post-processing is often necessary. For instance, in [148] the segment candidates were first selected by velocity zero crossing, and then the final breakpoints were identified from the segment candidates using an HMM.

The second class of approaches employs machine learning methodology to discover latent patterns from template libraries. HMM is often selected for segmentation of movement data, where each segment is treated as a hidden state, and the Viterbi algorithm is used to recover the state sequence [154]. Using regression-based techniques, a piecewise linear function was applied to fit the template data, and segmentation was performed when the difference between the data and the regression line was greater than a given threshold [155]. Traditional classifier methods (such as SVM) were also used for movement segmentation. In [148], all data points of motion sequences were assumed to be either segment points or non-segment points, and a trained SVM model was utilized to classify the points and segment the motion data.

## 7. Scoring Functions

Scoring functions are often used to convert the output values of movement evaluation algorithms to a meaningful performance score limited within a certain range [33], [37], [61], [62], [72], [126], [128], [138], [156], [157]. Concretely, it is important that the approaches for movement evaluation generate quality scores that are understandable and interpretable both by patients and medical professionals. For instance, quality scores that range from 0 to 100, or from 0 to 1, are easy to understand, record, and compare. On the other hand, the outputs of the quantitative algorithms described in Section 5 may be spread within a small or a large range of values, or they may even be negative numbers. Scoring functions are mathematical functions that map the outputs of the various approaches for movement evaluation to a convenient range of values that is meaningful to the end-users, and are therefore an important component of these systems. Scoring functions are also central to the validation and comparison of different approaches. Let $X = (x_1, x_2, \cdots, x_m)$ and $Y = (y_1, y_2, \cdots, y_n)$ denote the template and candidate training sequence, respectively. In [156], the dissimilarity between two motion sequences was converted to a performance score using the following formula: $S(X,Y) = 1 - \frac{d_{DTW}(X,Y)}{maximum\ distance}$, where *maximum distance* is the DTW distance between the reference movement and the practice movement captured when the user was not moving at all. Zhang et al. [138] proposed a more complex score function based on constructed lower and upper bounds of the DTW distance. Concretely, the score function is $S(X,Y) = 1 - \frac{d_{dtw}(X,Y) - d_{lb}(X,Y)}{d_{ub}(X,Y) - d_{lb}(xX,Y)}$, with $d_{lb}(X,Y) = \max \begin{cases} |\text{First}(X) - \text{First}(Y)| \\ |\text{Last}(X) - \text{Last}(Y)| \\ |\max(X) - \max(Y)| \\ |\min(X) - \min(Y)| \end{cases}$ and $d_{ub}(X,Y) = \max(m,n) \cdot \max \begin{cases} |\max(X) - \min(Y)| \\ |\min(X) - \max(Y)| \end{cases}$. In [128], the performance score function was defined as $S(X,Y) = 1 - \frac{d_{DTW}(X,Y)}{90 \times 8 \times s}$, where $s$ is the length of the optimal path and 8 represents the eight bone vectors extracted from raw motion data.

A score function on the basis of membership functions was introduced in [72]. Specifically, the score function is given by $S(X,Y) = \frac{1}{T}\sum_{t=1}^{T} P_{score}(t)$, with $P_{score}(t) = \sum_{i=1}^{N} r_i f_i(x_i(t) - y_i(t))$, $\sum_{i=1}^{n} r_i = 1$, where $N$ is the number of features, $T$ is the length of motion sequences $X$ and $Y$ which have been aligned using DTW, and $f_i$ is a Gaussian-shape function. In [61] and [62], a score function was derived using the log-likelihood of a trained HSMM. The total score for the $i^{th}$ subject is $score_i = $ $(CE_{max} - CE_{min}) \times \frac{\log L_i - \log L_{min}}{\log L_{max} - \log L_{min}} + CE_{min}$, where $CE_{max}, CE_{min}$ are the maximum and minimum scores of clinical evaluation and $\log L_{max}, \log L_{min}$ are maximum and minimum values of the log-likelihood.

In contrast to these template-based score functions, Capecci et al. introduced the following two target-based score functions in [33] and [126], respectively: $score(input) = \begin{cases} 0 & if\ \Delta - |target - input| < 0 \\ \frac{30}{\Delta}(\Delta - |target - input|) & otherwise \end{cases}$, and $score(input) = \left(1 + \left|\frac{input - target}{\Delta}\right|\right)^{-1}$, where *input* denotes the extracted features, the *target* is the designed clinical goal to achieve and $\Delta$ stands for the admitted tolerance. Different from the above scoring functions, Jung et al. [37] designed a function on the basis of classifiers. The reaching evaluation score of a session $i$ was defined as $score^i = \left[\sum_{f=1}^{8} O_{Lf}^i, \sum_{f=1}^{8} O_{Rf}^i\right]$, where $O_{Lf}^i$ and $O_{Rf}^i$ are the sub-classifier outputs for the left and right sides, respectively. The maximal score of [8, 8] corresponds to the subjects performed the experimental task perfectly with either the left or right side, whereas the minimal score of [1, 1] indicates the opposite.

The choice of a scoring function depends on the method used for movement evaluation. Researchers have paid less attention to the scoring function component of the systems for rehabilitation evaluation, and there is a lack of studies that perform a comprehensive evaluation of the various scoring functions introduced in related works.

## 8. Future Directions

This review summarized the motivations for automated rehabilitation evaluation, reviewed the main sensors for capturing human motion, and discussed existing evaluation approaches in the literature. In this section, several future research directions are discussed.

*Definition of movement features for automated progress evaluation*—In clinical practice, the quality of patient movements or exercise performance is often assessed subjectively based on visual observation by a clinician. Although some patient scenarios allow for the use of quantitative functional measures (e.g., postural sway analysis, reaction time, range of motion testing) or patient-reported questionnaires for evaluation, clinicians often rely on more subjective tests (e.g., Functional Movement Screen™, specific skill evaluation, muscle tests) or their individual understanding of the correct performance of an exercise without the use of a checklist, rubric, or strict rules to inform evaluation of patient function or performance [25], [28], [158]. The use of subjective and intuitive evaluation methods creates the risk of measurement errors due to outside factors (e.g., clinician bias, measurement imprecision, etc.) which affect the reliability and validity of the evaluation, and clinicians may overestimate patient performance or function [25], [159]. Defining sets of recommended criteria for movement evaluation may reduce the variability and subjectivity in related tests [25], [158], as well it can benefit the approaches for automated evaluation of exercises.

Although several research works have addressed this problem, the focus has been solely on quantifying rehabilitation movements for static body postures [49] and balance tests [81]. A recent study by Capecci et al. [62] proposed rules for assessing rehabilitation exercises based on attained target joint angles, target joint velocities, and postural constraints in accomplishing the goals of each exercise. However, the study proposed guidelines for only five selected rehabilitation exercises. The development of similar standard features for quantifying the various exercises that are commonly used in rehabilitation applications, and subsequently, the availability of databases with clinician-scored exercises, would provide the needed ground truth and facilitate the development of systems for automated performance monitoring and evaluation. Furthermore, valuation of the practical relevance of such approaches should also include non-inferiority studies, to provide a better understanding of whether unsupervised rehabilitation with systems for automated performance evaluation produces similar therapeutic benefits as clinician-supervised rehabilitation.

Such studies will furnish the necessary insights underpinning the usability and benefits of the systems for clinical support of at-home rehabilitation.

*Deep neural networks for feature learning*—The majority of prior related studies are based on manual selection of important movement features, or engineering new features from captured motion data. Such approaches have limitations since they require specialized expertise and motion kinematics knowledge to manually extract practical features from motion data, and the extracted features cannot be reused for new exercises. Although traditional feature engineering algorithms, such as PCA [47] and manifold learning [41], can be applied to encode local or global features for exercise evaluation, these methods generally assume certain preconditions. Deep learning-based approaches are widely used across various applications for encoding feature representations without the need for domain-specific knowledge. The ability of deep learning models to encapsulate highly nonlinear relations among sets of observed and latent variables, as well as the capacity to encode data features at multiple hierarchical levels of abstraction make them an attractive means for motion modeling and analysis. For example, Vakanski et al. [39] employed an autoencoder neural network for dimensionality reduction of rehabilitation data. Other related research proposed different architectures of neural networks for learning spatial and temporal features from movement data [160]–[164]. In [162], a deep recurrent network was designed to learn co-occurrence features from skeletal data through a novel regularization scheme. Similarly, Song et al. [163] proposed a deep network that introduced spatial and temporal attention subnetworks. The spatial attention mechanism facilitated the selection of dominant joints, whereas the temporal attention assigned greater weights to salient time frames. Despite the large body of previous literature and research on deep learning for motion modeling (e.g., for motion recognition, classification), little research has been conducted on movement evaluation in rehabilitation exercises.

*Large-scale rehabilitation datasets*—One of the reasons for the limited research on deep learning for evaluation of human motions in physical rehabilitation applications is the lack of large-scale annotated datasets of rehabilitation exercises. The publicly available datasets for rehabilitation evaluation covered in this review are of relatively small size, and typically do not offer clinically relevant scores provided by experienced medical professionals. The emergence of larger and more comprehensive datasets, such as UI-PRMD and KIMORE, provides a basis for research in this direction and optimism that the community will put efforts into the collection of new datasets.

*Combining movement quality with pain level evaluation*—Patients' pain level during a rehabilitation exercise session can reflect their health status, and thus, it can be an important indicator of the treatment outcome. Accordingly, a great deal of study focused on assessing pain level [109], [110], [112], [165], [166]. E.g., in [110] facial images captured by a smartphone were used to estimate the pain level of cancer patients. Milton et al. [165] studied the relation between common symptoms and health aspects, and found that the pain intensity produces stronger relationships when compared to other symptoms. Aung et al. [166] reviewed the literature on non-verbal expression of chronic pain to select factors that contribute to the occurrence of pain-related behaviors, and further discussed how the detection of pain-related behaviors could support rehabilitation. Integrating movement evaluation based on captured motion data and pain evaluation based on visual facial expressions into a single comprehensive rehabilitation indicator can be an exciting research direction.

*Fusing data from depth sensors and inertial sensors*—As we stated earlier, the extraction of skeletal data is a trivial task with existing vision/depth sensors. However, these sensors are susceptible to the external environment (such as occlusion of body parts and lighting conditions) and have limitations in providing accurate information. For example, Destelle et al. [167] improved the accuracy of skeletal data extracted from depth cameras with the use of inertial data. Chen et al. [168] reported enhanced movement recognition when fusing depth and inertial data. Moreover, the fusion of depth, skeleton, and inertial data for human action recognition has been explored in numerous other studies [169]–[173]. In contrast, data fusion from different types of sensors is rarely applied to assessing rehabilitation exercises, and it can be a promising research avenue.

*Motion capture using a smartphone*—At the present time, smartphones are accessible to most people as they are becoming more affordable. Most smartphones possess high-resolution cameras and advanced inertial sensors. Therefore, the use of smartphones to record human motion for exercise evaluation is appealing. Some efforts have been made in this regard [174]–[181]; however, these works have been limited to the use of inertial data for rehabilitation evaluation. In addition, there is no systematic study on the accuracy and reliability of these sensors for motion capture. In 2018, Apple introduced the TrueDepth IR (infrared) camera to the iPhone X line. The camera allows motion capture with the smartphone, which offers great potential for related medical applications. It is very likely that most of the other smartphone manufacturers will introduce similar motion capture technology to their models in the very near future. The use of smartphones for movement evaluation is particularly attractive and suitable for home-based rehabilitation.

*Combining evaluation with voice assistants*—Providing a qualitative or quantitative evaluation score of rehabilitation exercises to patients is far from sufficient to support effective implementation of at-home rehabilitation programs. The integration of voice assistants for conveying evaluation feedback to the patients can greatly improve the efficiency and user-friendliness of these types of systems. For instance, an integrated voice assistant (similar to Alexa or Google's voice assistant) can instruct the patient on the sequence of movements to perform or the correctness of the posture during a practice session, as well as provide suggestions on how to improve the exercise quality or which aspects of the movements are not performed correctly.

## 9. Conclusion

This paper presents a review of computational approaches for automatic evaluation of patient performance in rehabilitation exercise programs, with a focus on machine learning methods for quantification of the quality of patient movements performed in a home-based setting.

The review categorizes the pertinent approaches into three major groups: discrete movement score, rule-based, and template-based approaches. The main characteristics, advantages and disadvantages of these groups of approaches, and representative studies of related works in the literature are described in the paper. We also detail the sensory systems used for data collection of rehabilitation movements and provide a description of the respective datasets for rehabilitation evaluation. The study reviews respective publications on the related topics on feature engineering, movement segmentation, and scoring functions. Lastly, we list several recommendations for future directions in rehabilitation evaluation.

The advances in machine learning and the advent of inexpensive and reliable motion capture sensors have inspired an increased interest in automated evaluation of rehabilitation exercises. Related studies in the literature corroborated the feasibility and viability of such technology, and advocated that it can create substantial benefits both for patients and healthcare systems. Numerous research works reported high accuracy in predicting the level of correctness of patient performance in comparison to reference movement data collected with healthy subjects. In addition, comparative studies that employed clinicians' evaluation of movement quality as ground truth for validation of the computational approaches for exercise evaluation reported a high correlation in the assigned quality scores. These findings have been encouraging and evinced the potential of computational approaches for automated evaluation of rehabilitation exercises.

Despite the progress, there are still open questions and numerous challenges to overcome before we can witness a broad deployment of these systems in home-based and in-clinic settings. On one hand, modeling human movements remains a challenging problem and requires devising novel models that can successfully encapsulate the inherent variability in human movements. Furthermore, little research has been conducted on evaluating the impact of the approaches for rehabilitation evaluation on the long-term patient outcomes and whether the provided evaluation produces similar therapeutic benefits as clinician-supervised exercise programs. Another major impediment is the reliance of the greatest majority of the related systems on Kinect v1 or v2 sensors, whose production was discontinued by Microsoft in 2018. Fortunately, Microsoft introduced a new version Azure Kinect in 2019; however, the new sensor uses a different programming platform, and cannot re-use the programs developed for the older Kinect sensors in numerous research studies.

The recent introduction of cell-phones with motion tracking capabilities combined with the progress in cloud computing services offer enormous potential for widespread use of this technology. We believe that these systems will be ubiquitous in the near feature and they will play an important role in complementing the traditional approaches for evaluation of rehabilitation exercises.

## 10. Acknowledgments

This work was supported by the Institute for Modeling Collaboration and Innovation (IMCI) at the University of Idaho through NIH Award #P20GM104420.